\documentclass{article}

\usepackage[final]{neurips_2024}

\usepackage[utf8]{inputenc} 
\usepackage[T1]{fontenc}    
\usepackage{hyperref}       
\usepackage{url}            
\usepackage{booktabs}       
\usepackage{amsfonts}       
\usepackage{nicefrac}       
\usepackage{microtype}      
\usepackage{xcolor}         
\usepackage{graphicx}
\usepackage{wrapfig}
\usepackage{float}
\usepackage{arydshln}
\usepackage{amsmath}
\usepackage{adjustbox}
\usepackage{subcaption}
\usepackage[linesnumbered,boxed,commentsnumbered,ruled]{algorithm2e}
\usepackage{svg}
\newcommand{\Checkmark}{\checkmark}
\newcommand{\etal}{\textit{et al} }
\newcommand{\ie}{\textit{i}.\textit{e}. }
\newcommand{\eg}{\textit{e}.\textit{g}. }

\title{Not Just Object, But State: Compositional Incremental Learning without Forgetting
}

\author{%
  Yanyi Zhang~$^{1}$, Binglin Qiu~$^{1}$, Qi Jia~$^{1}$, Yu Liu~\thanks{corresponding author}~$^{1}$, Ran He~$^{2}$ \\
  ~$^{1}$~International School of Information Science \& Engineering, Dalian University of Technology\\
  $^{2}$~MAIS\&CRIPAC, Institute of Automation, Chinese Academy of Sciences\\
  \texttt{yanyi.zhang@mail.dlut.edu.cn, m1andy@mail.dlut.edu.cn}\\
  \texttt{jiaqi@dlut.edu.cn, liuyu8824@dlut.edu.cn, rhe@nlpr.ia.ac.cn}\\
}

\begin{document}
\maketitle

\begin{abstract}
Most incremental learners excessively prioritize coarse classes of objects
while neglecting various kinds of states (\eg color and material) attached to the objects. 
As a result, they are limited in the ability to reason fine-grained compositionality of state-object pairs. 
To remedy this limitation, we propose a novel task called \textbf{Compositional Incremental Learning} (composition-IL), 
enabling the model to recognize state-object compositions as a whole in an incremental learning fashion.
Since the lack of suitable benchmarks, we re-organize two existing datasets and make them tailored for composition-IL.
Then, we propose a prompt-based \textbf{Comp}osition \textbf{I}ncremental \textbf{L}earn\textbf{er} (\textbf{CompILer}),
to overcome the ambiguous composition boundary problem which challenges composition-IL largely. Specifically, we exploit multi-pool prompt learning, which is regularized by inter-pool prompt discrepancy and intra-pool prompt diversity.
Besides, we devise object-injected state prompting by using object prompts to guide the selection of state prompts.
Furthermore, we fuse the selected prompts by a generalized-mean strategy, 
to eliminate irrelevant information learned in the prompts.
Extensive experiments on two datasets exhibit state-of-the-art performance achieved by CompILer.
Code and datasets are available at: 
\url{https://github.com/Yanyi-Zhang/CompILer}.
\end{abstract}

\section{Introduction}
Class Incremental Learning (class-IL)~\cite{CL_survey_1,CL_survey_2,LwF,EWC} gathers increasing attention 
due to its ability to make the models learn new tasks rapidly, without forgetting previously acquired knowledge. 
Yet, traditional class-IL sets a strict limit on the old classes such that they should not recur in newly incoming tasks.
To break such a strict limitation, recent studies develop a new setting mostly called Blurry Incremental Learning (blur-IL)~\cite{MVP,CLIB},
where the incremental sessions allow the recurrence of previous classes, resulting in a more realistic and flexible scenario.
Despite such empirical progresses on incremental learning, they aim to improve object classification only, 
while overlooking fine-grained states attached to the objects.
For instance, analyzing how the clothing styles (akin to states) 
have changed over time is important for forecasting the future trends that will emerge.



\begin{figure*}[!t]
  \centering
  \includegraphics[width=\textwidth]{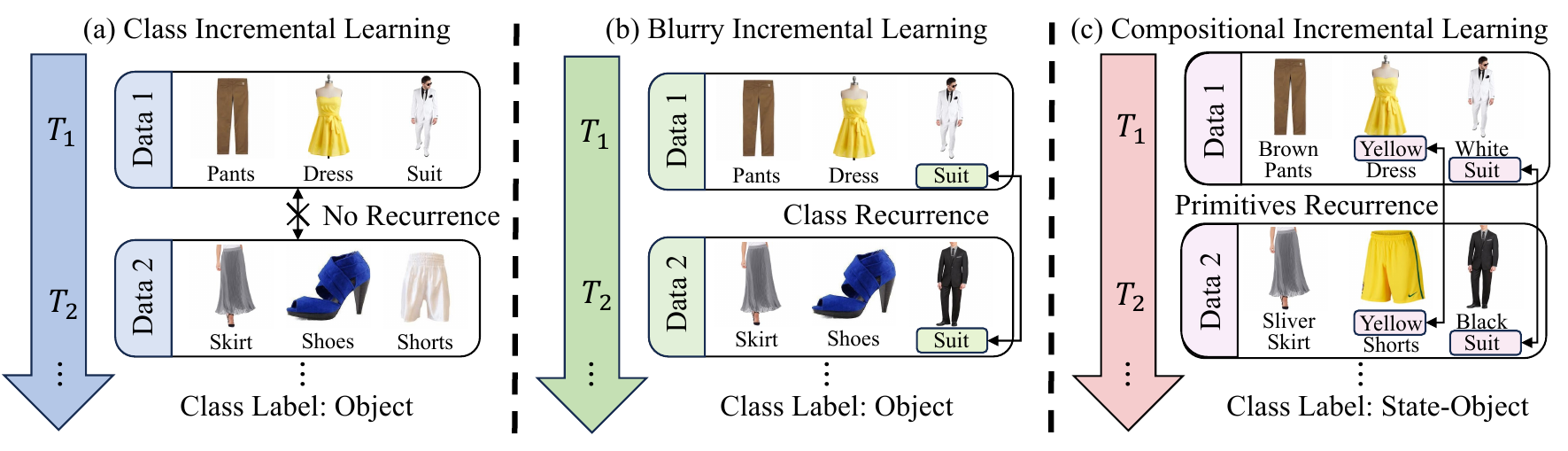}
  \vspace{-0.4cm}
   \caption{Differences between Class Incremental Learning (class-IL), Blurry Incremental Learning (blur-IL), and Compositional Incremental Learning (composition-IL). The object classes are not allowed to recur in the class-IL scenario, whereas they may recur randomly in the blur-IL scenario. Different from them, the classes in composition-IL involve state-object compositions apart from the object classes. Besides, the compositions do not reoccur, but the primitives (states or objects) may randomly reappear across incremental sessions.}
   \label{fig:Fig1}
\end{figure*}

To simultaneously model objects and their states, some efforts are dedicated to Compositional Learning whose aim is how to equip the models with \textit{compositionality}~\cite{CognitiveScience1,CognitiveScience2,CognitiveScience3}.
The core of compositional learning lies in the structure of class labels, which conceptualizes a state-object pair (\eg ``Brown Pants'' and ``Yellow Dress'') as a whole, rather than a lonely object label. In this way, the model can dissect and reassemble learned knowledge, 
achieving a more fine-grained understanding about the objects.
However, existing works are mainly focused on zero-shot generalization from seen compositions to unseen ones~\cite{AttrAsOpe,Task-driven,CZSL_AAAI_2024,CZSL_CVPR_2024}, whereas none of them consider the challenging fact that the model must deal with
a significantly larger number of composition classes than object classes.
As a result, it is hardly feasible to learn all compositions by training the model once.

To remedy the limitations inherent in incremental learning and compositional learning, 
we conceive a novel task named \textbf{Compositional Incremental Learning} (composition-IL),
enabling the model to continually learn new state-object compositions in an incremental fashion.
As compared in Fig.~\ref{fig:Fig1}, we can see that composition-IL integrates the characteristics of class-IL and blur-IL.
Although the composition classes are disjoint across incremental tasks, the primitive classes (\ie objects and states) 
encountered in old tasks are allowed to reappear in new tasks.
Unfortunately, existing incremental learning approaches are challenged by such a compositional scenario,
because their models excessively prioritize the object primitives while neglecting the state primitives.
Consequently, the compositions with the same object but with different states become ambiguous and indistinguishable.

To tackle the problem, we propose a rehearsal-free and prompt-based \textbf{Comp}ositional \textbf{I}ncremental \textbf{L}earn\textbf{er} (\textbf{CompILer}).
Specifically, our model comprises of three primary components: multi-pool prompt learning, object-injected state prompting, and generalized-mean prompt fusion. 
Firstly, we construct three prompt pools for learning the states, objects and compositions individually. 
Upon that, we add extra restrictions to regularize the inter-pool prompt discrepancy and intra-pool prompt diversity.
This multi-pool prompt learning paradigm strengthens the fine-grained understanding and reasoning towards primitive concepts and their compositions.
In addition, as the state classes are more difficult to distinguish than the object ones, we propose object-injected state prompting
which incorporates object prompts to guide the selection of state prompts.
Furthermore, we fuse the selected prompts by a generalized-mean fusion manner, 
which helps to adaptively eliminate irrelevant information learned in the prompts.
Last but not least, we also leverage symmetric cross-entropy loss to alleviate the impact of noisy data during training.

In summary, the main contributions in this work are encapsulated as follows:
(1) We devise a new task coined compositional incremental learning (composition-IL). 
It enables learning fine-grained state-object compositions continually while the isolated primitive concepts can randomly recur in incremental tasks.
(2) To address the lack of datasets, we re-organize two existing datasets such that they are tailored specifically for composition-IL.
For the two new datasets (Split-Clothing and Split-UT-Zappos), we split them into 5 and 10 incremental tasks for evaluating the methods.
(3) We propose a novel learning-to-prompt model for composition-IL, namely CompILer. 
Our state-of-the-art results on Split-Clothing and Split-UT-Zappos validate the effectiveness of CompILer
for incrementally learning new compositions without forgetting old ones.

\section{Related Work}
\textbf{Incremental Learning.}
The approaches to addressing catastrophic forgetting for incremental learning can be broadly grouped into four categories: regularization based methods~\cite{IL_regular_1,IL_regular_2} aim to protect influential weights of old experiences from updating; knowledge distillation based methods~\cite{LwF,IL_distill_1} distill knowledge from the model trained on the previous tasks and adapt it to new tasks; rehearsal based methods~\cite{icarl,IL_replay_1,IL_replay_2} 
require a memory buffer to store some old data, so as to make the network remember previous tasks;
parameter isolation methods allocates different model parameters to each task, to prevent any possible interference.
Different from the methods, L2P~\cite{L2P} proposes an innovative learning-to-prompt paradigm, 
which incorporates plasticity and stability through adapting a set of learnable prompt tokens on top of a frozen pre-trained backbone.
Inspired by L2P~\cite{L2P}, more recent works~\cite{Dual,HiDe,Consistent_Prompting,Prompt_Gradient_Projectio} take full advantage of
various prompt tuning strategies, achieving new state-of-the-art performance for incremental learning.
However, such methods take into account object classes solely, while neglecting various kinds of state classes associated with the objects.
To this end, our work proposes compositional incremental learning with the purpose to continually identifying the composition classes of state-object pairs.
Note that, Liao,~\etal~\cite{CLC} conduct an initial study toward the compositionality in incremental learning, 
whereas their attention is on the composition of multiple object classes (\eg ``Car'' and ``Person'') in one image, rather than the state-object compositions in this work.

\textbf{Compositional Learning.}
A major line of compositional learning research focuses on Compositional Zero-Shot Learning (CZSL)~\cite{RedWine},
which aims to infer unseen state-object compositions by acquiring knowledge from seen ones. 
Subsequent approaches building upon the CZSL setting further incorporate graph neural networks to 
model the dependency between primitives and compositions~\cite{CGE}, and employ cosine classifiers to avoid being overly biased toward seen compositions~\cite{CompCos}.
Other approaches~\cite{symnet,symnet-pami,SCEN} propose training two classifiers to identify states and objects separately. 
The latest works~\cite{PMGNet,ADE,CANet,CoT,CSCNet} model both composition and primitives simultaneously, achieving state-of-the-art results. 
Albeit the numerous attempts made in compositional learning, they fail to consider an incremental learning paradigm given
the increasing number of composition classes in open-world scenarios. 
Besides, directly applying CZSL methods to composition-IL might lead to a stale and decaying performance on forgetting. 
By contrast, our proposed CompILer markedly bypasses catastrophic forgetting with the help of multi-pool prompt learning.


\begin{figure*}[t]
  \centering
  \includegraphics[width=0.95\textwidth]{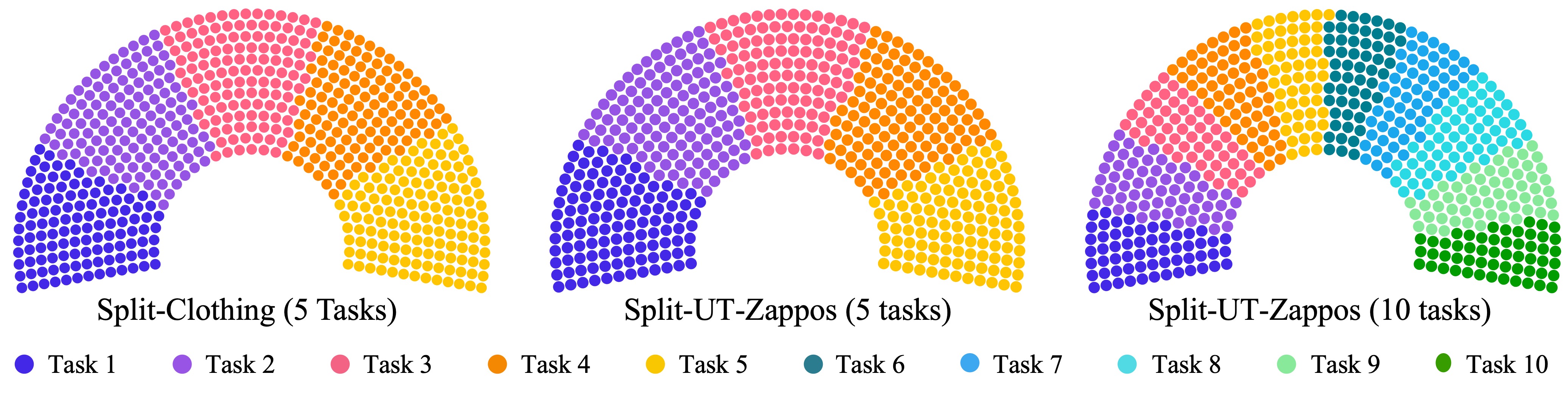}
   \vspace{-0.1cm}
   \caption{Data Statistics of Split-Clothing and Split-UT-Zappos for tasking composition-IL. Split-Clothing is divided into a 5-task scenario, while Split-UT-Zappos includes both 5-task and 10-task scenarios. In all settings, the number of images per task has been balanced properly.
   }
   \label{parliament}
\end{figure*}

\section{Preliminaries}
In this section, we firstly define the task of composition-IL, and then introduce two datasets we construct for the task, followed by revealing the ambiguous composition boundary problem.
\subsection{Problem Definition}
For composition-IL, a model sequentially learns $N$ tasks $\mathcal{T} =\left \{ \mathcal{T}_1, \mathcal{T}_2,\cdots \mathcal{T}_N \right \}$ corresponding to a set of composition classes $\mathcal{C} =\left \{ \mathcal{C}_1, \mathcal{C}_2,\cdots \mathcal{C}_N \right \}$. 
We note that the composition classes between incremental tasks are always disjoint, which means $\mathcal{C}_i\cap \mathcal{C}_j=\emptyset$ for any $i \ne j$.
Different from the composition classes, the primitive classes are allowed to recur in different tasks.
That means it allows the tasks to share some primitive concepts of objects and states.
Therefore, we can define the set of all state and object classes with $\mathcal{S}=\left\{s_1, s_2, \cdots, s_n\right\}$ and $\mathcal{O}=\left\{o_1, o_2, \cdots, o_m\right\}$, respectively. Given each image $x$, it has a composition label $c$ which is constructed with a state label $s$ and an object label $o$, \ie $c = <s, o>$, where $c \in \mathcal{C}$, $s \in \mathcal{S}$ and $o \in \mathcal{O}$. 
We take the example of ``red shirt'', where ``red'' is denoted with $s$, ``shirt'' corresponds to $o$, and ``red shirt'' is expressed with $c$. 


\subsection{Dataset Construction}
\label{subsection: Dataset Construction}
As there are no existing datasets suitable for composition-IL, 
we re-organize the data in Clothing16K~\cite{IVR} and UT-Zappos50K~\cite{ut-zappos}, 
and construct two new datasets tailored for composition-IL, namely \textbf{Split-Clothing} and \textbf{Split-UT-Zappos}. 
To be more specific, we firstly sort the composition classes based on the number of their images, 
and then select the foremost 35 compositions from Clothing16K and the top 80 from UT-Zappos50K,
so as to construct Split-Clothing and Split-UT-Zappos, respectively.
In this way, Split-Clothing encompasses 9 states and 8 objects while Split-UT-Zappos consists of 15 states and 12 objects in total.
For Split-Clothing, we randomly partition the compositions into 5 tasks. 
Regarding Split-UT-Zappos, the compositions are sorted by count and are evenly divided into 5 and 10 tasks.
The image distribution for each task is shown in Fig.~\ref{parliament}.
Note that, we elaborate more details on both datasets in the following technical appendix.


\begin{figure*}[t]
    \centering
    \includegraphics[width=0.9\textwidth]{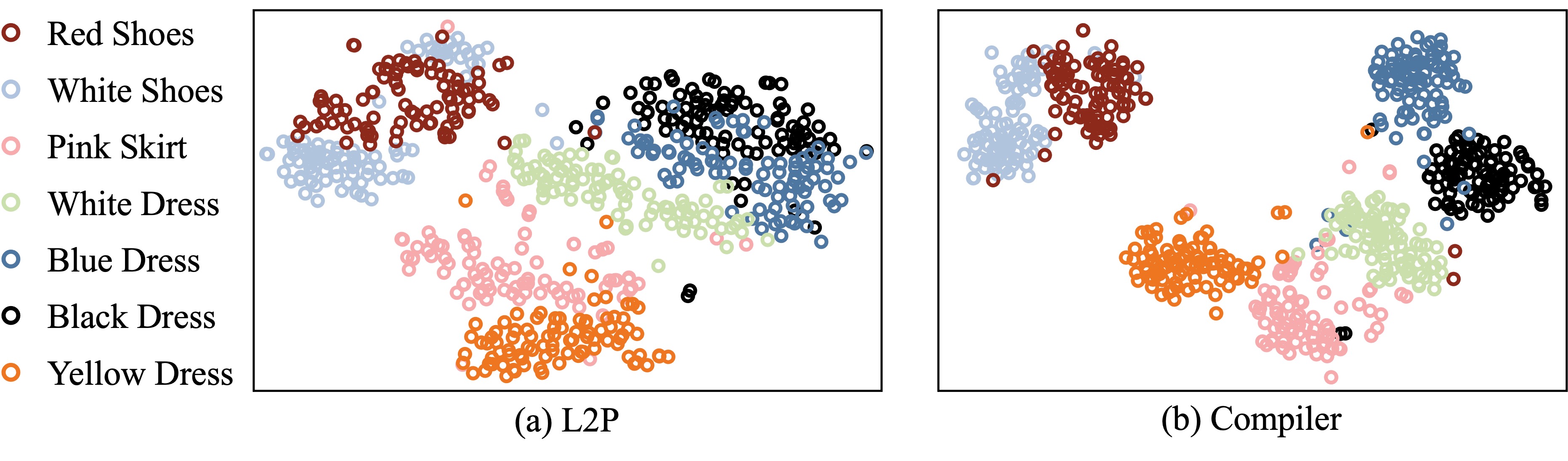}
    \vspace{-0.2cm}
    \caption{\textit{t}-SNE feature distributions of seven compositions from the Split-Clothing benchmark. 
    For the compositions with the same object but with different states, our CompILer achieves more distinguishable boundaries than the L2P baseline.}
    \label{fig:tsne}
\end{figure*}
\subsection{Revealing the Ambiguous Composition Boundary}
The main stumbling block in composition-IL is the ambiguous composition boundary. 
Although the composition label consists of two primitives (\ie object and state), 
we note that the model excessively prioritizes the object primitive while neglecting the state primitive. 
Consequently, the compositions with the same object but with different states become ambiguous and indistinguishable. 
To prove that, we apply L2P~\cite{L2P} to composition-IL, whereas it is challenged by significant ambiguities in composition classification. 
As illustrated in Fig.~\ref{fig:tsne} (a), the \textit{t}-SNE visualization showcases the entanglement among the compositions 
like ``white dress'', ``black dress'' and ``blue dress''.
We conjecture that this ambiguous problem tends to become more severe when more tasks are arriving incrementally.
To address it, we propose a new model namely CompILer, 
which disentangles compositions and primitives via a multi-pool prompt learning. 
Advantageously, our method promotes the learning on the states and establishes clearer composition boundaries, as shown in Fig.~\ref{fig:tsne} (b). 

\section{Methodology}
\begin{figure*}[t]
  \centering
  \includegraphics[width=\textwidth]{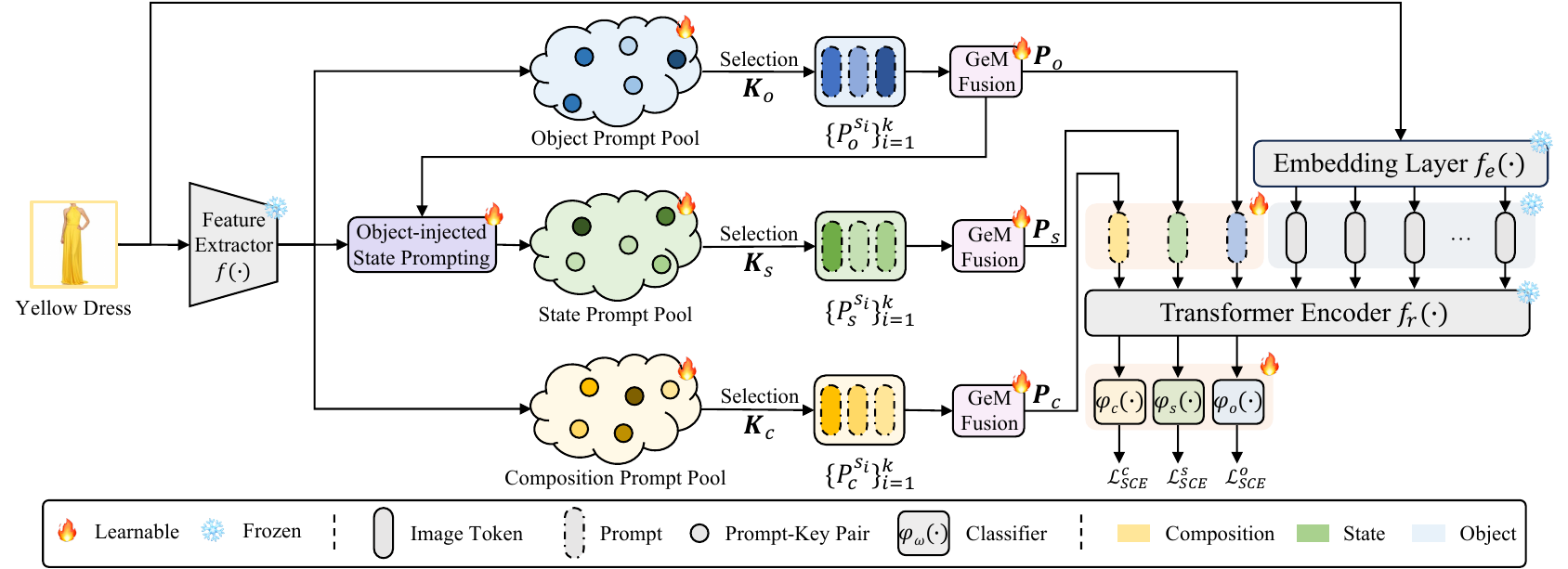}
   \vspace{-0.2cm}
   \caption{Overall architecture of our composition incremental learner (CompILer), which comprises multi-pool prompt learning, object-injected state prompting, and generalized-mean prompt fusion. The multi-pool prompt learning mechanism captures information related to states, objects, and their compositions, each through a dedicated pool. The object-injected state prompting utilizes the object prompt to promote the state representation learning. Moreover, the generalized-mean prompt fusion is used to prioritize the useful prompts and diminish the irrelevant ones.}
   \label{fig:overview}
\end{figure*}

\textbf{Overview.}
We leverage the learning-to-prompt paradigm~\cite{L2P} and develop a novel compositional incremental learner (CompILer) tailored specifically for composition-IL. 
As depicted in Fig.~\ref{fig:overview}, CompILer comprises three primary components: multi-pool prompt learning, object-guided state prompting, 
and generalized-mean prompt fusion. Firstly, we initialize three prompt pools dedicated to learning and storing visual information 
related to states, objects and their compositions. 
In order to differentiate the knowledge learned across and within prompt pools, 
we define inter-pool discrepant loss and intra-pool diversified loss jointly. 
We then employ object prompts to guide the selection of state prompts, thereby improving the state representation learning. 
Moreover, we utilize a generalized-mean fusion to integrate the selected prompts in a learnable manner. 
Ultimately, we optimize the classification objective with symmetric cross-entropy loss, to alleviate the effect of noisy data.

\subsection{Multi-pool Prompt Learning}
\label{Multi-pool Prompt Learning}
The learning-to-prompt paradigm~\cite{L2P, NotMeetStrongPT}, especially suitable for large pre-trained backbones, 
has opened up a new path for incremental learning.
It has proven to incorporate plasticity and stability better through adapting a set of learnable tokens in a prompt pool 
to a frozen pre-trained backbone. 
Nevertheless, existing prompt-based approaches are initially designed for class-IL, thereby building a single prompt pool for object classification solely.
when dealing with state-object composition classification, they tend to excessively prioritize the object primitive while neglecting the state primitive.
To this end, we propose to construct three discrepant and diversified prompt pools $\mathbb{P}_s$, $\mathbb{P}_o$ and $\mathbb{P}_c$, 
which serve to learn visual information related to states, objects and their compositions, respectively. 
Besides, each pool is associated with a set of learnable keys $\mathbb{K}_{\omega}$ for query-key prompt selection. The three prompt pools and their keys are defined as:
\begin{equation}
\mathbb{P}_{\omega} = \left \{ P_{\omega}^1,P_{\omega}^2,\cdots P_{\omega}^M \right \},
\mathbb{K}_{\omega} = \left \{ K_{\omega}^1,K_{\omega}^2,\cdots K_{\omega}^M \right \},
\omega \in \left \{ s,o,c \right \},
\end{equation}
where $P_{\omega}^i\in \mathbb{R}^{L\times D}$ is a single prompt with token length $L$ and embedding dimension $D$. $K_{\omega}^i \in \mathbb{R}^{D}$, the key of $P_{\omega}^i$, is a learnable token with the same size.
$M$ is the number of prompts in each pool. 

One important concern in such multi-pool prompt learning is how to 
enrich the prompts with the avoidance of identical pools.
To achieve it, we consider integrating \textbf{inter-pool prompt discrepancy} and \textbf{intra-pool prompt diversity} jointly.
On the one hand, the inter-pool prompts should be discrepant as the visual information about states, objects, and compositions should be different. 
One the other hand, within each pool, the intra-pool prompts should be diversified so to capture more comprehensive features from all the classes.

In practice, we formulate a unified objective to regularize both inter-pool discrepancy and intra-pool diversity, by leveraging a simple and effective directional decoupled loss used in~\cite{PIP}. 
The directional decoupled (dd) loss between any two pools (\eg $P_{i}$ and $P_{j}$) is formulated as:
\begin{equation}
\mathcal{L}_{dd}^{(i,j)}=\frac{2}{M(M-1)} \sum_{n=1}^{M} \sum_{m=1}^{M} \max \left(0, \theta_{\text {thre}}-\theta_{n m}\right), 
\end{equation}
\begin{equation}
\theta_{n m}=\cos ^{-1}\Big(\frac{(P_{i}^{n})^{\mathrm{T}} P_{j}^{m}}{\max (\|P_{i}^{n}\|_{2},\epsilon)\cdot  
\max(\|P_{j}^{m}\|_{2}, \epsilon)}\Big),
\end{equation}
where $\theta_{nm}$ measures the angle between any two prompts, $n$ and $m$;
$\epsilon$ is a scalar to avoid division by zero.
Note that, $\mathcal{L}_{dd}^{(i,j)}$ encourages the angles between each prompt to be at least $\theta_{\text{thre}}$ degrees.
Since $(i,j)$ is unordered Cartesian product of $\omega$, \ie$(i,j) \in \left \{ (i,j)\mid i\in \omega \wedge j\in \omega  \right \} $,
the inter-pool prompt discrepancy loss for the three pools can be expressed with $\mathcal{L}_{inter} = \mathcal{L}_{dd}^{(s,o)} + \mathcal{L}_{dd}^{(s,c)} + \mathcal{L}_{dd}^{(o,c)}$, and the intra-pool prompt diversity loss becomes $\mathcal{L}_{intra} = \mathcal{L}_{dd}^{(s,s)} + \mathcal{L}_{dd}^{(o,o)} + \mathcal{L}_{dd}^{(c,c)}$. 
As opposed to $\mathcal{L}_{inter}$, $\mathcal{L}_{intra}$ computes the angle between any two prompts within the same pool. Thus, it contains the case when $n=m$, for which we set $\theta_{\text {thre}}-\theta_{n m}=0$.

\clearpage

\subsection{Object-injected State Prompting}
\begin{wrapfigure}{R}{0.5\textwidth}
    \centering
    \includegraphics[width=0.5\textwidth]{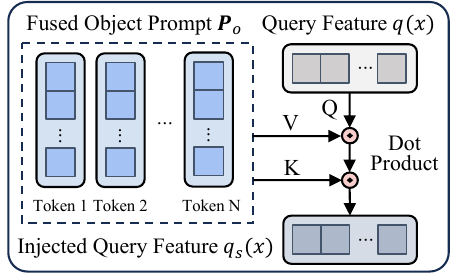}
    \vspace{-0.4cm}
    \caption{Architecture of object-injected state prompting. Query feature serves as Q, while fused object prompt serves as both K and V.}
    \label{fig:OBG}
\end{wrapfigure}

Akin to the query-key matching mechanism in other work~\cite{L2P,Dual,s-prompt}, 
we utilize a fixed feature extractor $f(\cdot)$ to obtain a query feature $q(x)=f(x)[0,:]$, determining which prompts in the pool to be selected.
However, pre-trained backbones are typically trained for object classification, thus under-performing for state representation learning.
In addition, it is more difficult to predict the state classes due to their more abstract and fine-grained characteristics.
To tackle this problem, we strategically inject object prompts to guide the selection of state prompts.
Intuitively, once we have learned knowledge about the object class, it may be easier to predict the correct state class and 
avoid mistaken results. 
For instance, given an object is ``heels'', we can expect that the corresponding state is unlikely to be ``canvas'' or ``plastic''.
To summarize, we select object and composition prompts in each pool 
based on the original query feature, which means $q_{o}(x)=q_{c}(x)=q(x)$;
but for the selection of state prompts, we propose object-injected state prompting to ameliorate the query feature as shown in Fig.~\ref{fig:OBG}.
 
Specifically, we employ the fused object prompt $\boldsymbol{P}_o$ (see Sec.~\ref{subsection:GeM}) to perform cross attention on the query feature $q(x)$, resulting in object-injected query feature $q_{s}(x)$ for the state prompt selection:
\begin{equation}
    q_{s}(x) = \text{CrossAttn}(q(x),\boldsymbol{P}_o)=\text{Softmax}(\frac{q(x)W^Q\cdot \boldsymbol{P}_oW^K}{\sqrt{D} } )\cdot \boldsymbol{P}_oW^V,
\end{equation}
where $W^Q$, $W^K$ and $W^V$ are learnable projections. 
To establish alignment between the query and the selected prompts,
we optimize a surrogate loss for state, object and composition prompting jointly:
\begin{equation}
\label{surrogate}
\mathcal{L}_{sur} = \sum_\omega  \sum_{q_{\omega}}\text{COS}(f_\omega(x), K_{\omega}^{s_i}),\
\omega \in \left \{ s,o,c \right \},
\end{equation}
where $\text{COS}(\cdot ,\cdot )$ denotes cosine similarity, $\mathbf{K}_{\omega}$ represents the subset of top-k keys selected from $\mathbb{K}_{\omega}$, and ${\left\{s_{i}\right\}_{i=1}^{k}}$ is a subset of top-k indices from $[1,M]$ (prompt number).
Despite the simplicity of the object-injected state prompting, it facilitates more judicious prompt selection within the state prompt pool, alleviating the hurdles posed by state learning.



\subsection{Generalized-mean Prompt Fusion}
\label{subsection:GeM}
After obtaining the selected top-k prompts ${\left\{P_{\omega}^{s_i}\right\}_{i=1}^{k}}$,
the next step is fusing these prompts into a single prompt.
It is general to utilize a simple mean pooling whereas it overlooks the relative importance of each prompt.
Besides, when the prompts contain information that is unrelated or contradictory to 
current task, it is critical to strengthen useful prompts and eliminate irrelevant ones.
To this end, we draw inspiration from generalized-mean pooling~\cite{GeM} 
and exploit generalized-mean (GeM) prompt fusion which is given by:
\begin{equation}
\label{eq gem}
\boldsymbol{P}_{\omega}=\text{GeM}_\omega(P_{\omega }^{s_1}, P_{\omega }^{s_2}, \cdots, P_{\omega }^{s_{k}})=\left(\frac{1}{k} \sum_{i=1}^{k} {P_{\omega}^{s_i}}^{\eta }\right)^{\frac{1}{\eta}},   \omega \in \left \{ s,o,c \right \},
\end{equation}
where $\eta$ is a learnable parameter. When $\eta=1$, GeM becomes mean pooling; as $\eta$ approaches infinity ($\eta\to\infty$), 
it converges to max pooling. 
By taking over mean and max pooling, GeM learns to achieve an optimal fusion, mitigating the influence of irrelevant information present in the prompts.

\subsection{Training and Inference}
\textbf{Classification Objective.}
We prepend three fused prompts (\ie $\boldsymbol{P}_{s}$, $\boldsymbol{P}_{o}$ and $\boldsymbol{P}_{c}$) with $x_e$, which is the output from a ViT embedding layer $f_e(\cdot)$.
The extended token sequence is $x_p = \left [ \boldsymbol{P}_{c}; \boldsymbol{P}_{s}; \boldsymbol{P}_{o}; x_e \right ]$.
Then, we feed $x_p$ to a transformer encoder layer $f_r(\cdot)$ and achieve $\boldsymbol{P}_{s}^r$, $\boldsymbol{P}_{o}^r$ and $\boldsymbol{P}_{c}^r$ for classifying state, object and composition classes, respectively.
We estimate the probability via a classifier $\varphi_\omega(\cdot)$:
$p(\omega \mid x) = \varphi_\omega(\boldsymbol{P}_{\omega}^r)$.
For each image $x$, we denote its ground-truth distribution over labels with $q(\omega \mid x)$. When $\omega$ is consistent with the ground truth, then $q(\omega \mid x) = 1$; otherwise, $q(\omega \mid x)=0$. As a result, the cross entropy (CE) loss used for classification objective is:
\begin{equation}
\mathcal{L}_{CE}^\omega = -\sum_{\omega=1}^{\Omega}q(\omega \mid x)\log{p(\omega \mid x)}, \Omega \in  \left [   \left | \mathcal{S} \right |,\left | \mathcal{O} \right |,\left | \mathcal{C} \right |\right ],
\end{equation}
where $\Omega$ represents the number of classes. 
However, the model optimized with a standard CE loss is easily
affected by noisy samples during training. 
Instead, we advocate using a symmetric cross entropy loss (SCE)~\cite{SCE},
which incorporates an additional term called reverse cross entropy (RCE),
to mitigate the impact of noisy data. 
Contrary to CE, the formula for RCE loss is defined as:
\begin{equation}
\mathcal{L}_{RCE}^\omega = -\sum_{\omega=1}^{\Omega}p(\omega \mid x)\log{q(\omega \mid x)}, \omega \in \left \{ s,o,c \right \}.
\end{equation}

Then, the SCE loss combines two loss terms by $\mathcal{L}_{SCE}^\omega = \mathcal{L}_{CE}^\omega + \alpha  \mathcal{L}_{RCE}^\omega$, where $\alpha$ is a hyper-parameter that controls the weight of the RCE term.
As a result, the whole SCE loss becomes $\mathcal{L}_{SCE} = \mathcal{L}_{SCE}^c+\beta (\mathcal{L}_{SCE}^s+\mathcal{L}_{SCE}^o)$, where $\beta$ adjusts the weights between primitives and compositions.

\textbf{Total Loss.}
The total loss for training the whole CompILer model is:
\begin{equation}
    \mathcal{L}_{total}=\lambda_1 \mathcal{L}_{inter}+\lambda_2 \mathcal{L}_{intra}+\lambda_3 \mathcal{L}_{sur}+\mathcal{L}_{SCE},
    \label{final_loss}
\end{equation}
where $\lambda_1$, $\lambda_2$, $\lambda_3$ are hyper-parameters balancing different terms.

\textbf{Inference.}
During inference, we incorporate the primitive probabilities 
to aid the composition probability. Hence, the final probability for composition classification is expressed with:
\begin{equation}
    p_{\text{final}}(c \mid x) = p(c \mid x) + \mu (p(s \mid x) + p(o \mid x)),
\end{equation}
where $\mu$ adjusts the probabilities.

\begin{table}[t]
\centering
\caption{Avg Acc and FTT results on Split-Clothing (5 tasks) and Split-UT-Zappos (5 and 10 tasks). The best results are marked in \textbf{bold}. All results with standard deviations are averaged over three runs.}
\resizebox{\linewidth}{!}{
\label{avg acc and ftt}
\begin{tabular}{l||cc||cc||cc}
\hline
Datasets & \multicolumn{2}{c||}{Split-Clothing (5 tasks)} & \multicolumn{2}{c||}{Split-UT-Zappos (5 tasks)} & \multicolumn{2}{c}{Split-UT-Zappos (10 tasks)}\\
\hline
Metrics & Avg Acc(↑) & FTT(↓) & Avg Acc(↑) & FTT(↓) & Avg Acc(↑) & FTT(↓)\\
\hline
Upper Bound & 97.02±0.10 & - & 68.71±0.41 & - & 68.71±0.41 & - \\
\hdashline
EWC~\cite{EWC} & 47.89±0.87 & 52.75±0.44 & 37.59±2.06 & 55.70±2.76 & 24.63±0.94 & 61.31±2.29 \\
LwF~\cite{LwF} & 49.96±0.68 & 44.22±0.53 & 40.15±0.43 & 49.61±0.68 & 30.38±1.41 & 58.15±0.20 \\
iCaRL~\cite{icarl} & 68.65±0.41 & 31.74±1.89 & 37.78±2.14 & 55.06±3.50 & 31.40±1.96 & 59.65±2.40 \\
\hdashline
L2P~\cite{L2P} & 80.22±0.41 & 14.23±0.44 & 42.20±2.18 & 20.41±2.76 & 31.65±0.16 & 31.02±1.62 \\
Deep L2P++\cite{L2P,CODA} & 80.55±0.45 & 12.60±1.90 & 42.37±0.65 & 30.10±1.56 & 30.68±0.35 & 32.20±1.96 \\
Dual-Prompt~\cite{Dual} & 87.87±0.63 & 7.71±0.25 & 43.30±0.19 & 19.41±2.80 & 33.01±1.65 & 24.61±1.11 \\
CODA-Prompt~\cite{CODA} & 86.35±0.20 & 8.99±0.71 & 43.35±0.29 & 21.76±2.45 & 31.40±0.36 & 30.54±2.63 \\
LGCL~\cite{Language_ICCV} & 87.32±0.10 & 7.58±0.06 & - & - & 33.56±0.31 & \textbf{24.37}±0.56 \\
\hdashline
\textbf{Sim-CompILer} & 88.38±0.08 & 8.01±0.42 & 45.70±0.68 & 20.06±0.62 & 33.30±0.10 & 30.31±0.03
\\
\textbf{CompILer} & \textbf{89.21}±0.24 & \textbf{7.26}±0.60 & \textbf{46.48}±0.26 & \textbf{19.27}±0.75 & \textbf{34.43}±0.07 & 28.69±0.82 \\
\hline
\end{tabular}
}
\end{table}

\begin{wraptable}{R}{0.6\textwidth}
\centering
\vspace{-0.35cm}
\caption{State, Object and HM results on Split-Clothing. The best results are marked in \textbf{bold}.}
\label{split-clothing}
\resizebox{0.6\textwidth}{!}{
\begin{tabular}{l||ccc}
\hline
Datasets & \multicolumn{3}{c}{Split-Clothing (5 tasks)} \\
\hline
Metrics & State & Object & HM \\
\hline
Upper Bound & 97.44±0.08 & 97.09±0.10 & 97.26±0.08 \\
\hdashline
EWC~\cite{EWC} & 86.49±0.97 & 52.72±1.30 & 67.50±0.97 \\
LwF~\cite{LwF} & 87.11±0.66 & 54.57±0.69 & 67.10±0.33 \\
iCaRL~\cite{icarl} & 91.21±1.05 & 71.70±0.99 & 80.28±0.74 \\
\hdashline
L2P~\cite{L2P} & 83.03±0.42 & 95.56±0.57 & 88.85±0.16 \\
Dual-Prompt~\cite{Dual} & 90.77±0.25 & 94.18±0.31 & 92.45±0.20 \\
LGCL~\cite{Language_ICCV} & 91.45±0.20 & 94.87±0.33 & 93.13±0.10 \\
\hdashline
\textbf{Sim-CompILer} & 91.15±0.10 & 96.32±0.02 & 93.66±0.02
\\
\textbf{CompILer} & \textbf{91.81}±0.23 & \textbf{96.67}±0.01 & \textbf{94.18}±0.06 \\
\hline
\end{tabular}
}
\label{tab: hm in clothing}
\end{wraptable}

\section{Experiments}
\subsection{Datasets and Metrics}
We conduct experiments on two newly split datasets: Split-Clothing and Split-UT-Zappos as elucidated in Section~\ref{subsection: Dataset Construction}.
We assess the overall performance on compositions using Average Accuracy (\textbf{Avg Acc}) and Forgetting (\textbf{FTT}). A higher Avg Acc signifies stronger recognition abilities, while a lower FTT indicates improved resilience against forgetting. Additionally, we provide individual Average Accuracy scores on states and objects, denoted as \textbf{State} and \textbf{Object} for simplicity. These metrics imply the ability to recognize fine-grained primitives. Furthermore, we calculate the Harmonic Mean (\textbf{HM}) between State and Object, \ie $ HM = 2\times \frac{(State\times Object)}{(State + Object)} $.
We provide more emphasis to Avg Acc and HM due to their more comprehensive assessment. Avg Acc encompasses the plasticity and stability~\cite{CODA,Consistent_Prompting} and HM provides a holistic evaluation on both state and object.


\subsection{Implementation Details}
For a fair comparison with previous works~\cite{L2P,Dual,Consistent_Prompting,CODA}, we also employ ViT B/16~\cite{ViT_B16} pretrained on the ImageNet 1K dataset as the feature extractor and backbone. 
For multi-pool prompt learning, the size of each pool is set to $20$, and each prompt has $5$ tokens.
We select top-5 prompts from each pool and generate a fused prompt.
During training, we utilize the Adam optimizer~\cite{adam} with a batch size of $16$. The whole CompILer undergoes training for $25$ epochs on the Split-Clothing, for $10$ epochs on the 5-task Split-UT-Zappos, and for $3$ epochs on the 10-task Split-UT-Zappos. For the Split-Clothing and the 10-task Split-UT-Zappos, we set the learning rate to $0.03$, while we use a learning rate of $0.02$ for the 5-task Split-UT-Zappos.
Note that, for all the methods, their results are averaged over \textbf{three runs} with the corresponding standard deviations reported to mitigate the influence of random factors.

As there are a few hyper-parameters in the model, we conduct a rigorous tuning on them. For instance, we set $\theta_{\text {thre}}$ to $\frac{\pi}{2}$ for all settings. 
For Split-Clothing, the loss weights $\lambda_1$ and $\lambda_3$ are set to $0.1$; $\lambda_2$ is set to $10^{-7}$; $\alpha$ and $\beta$ for SCE loss  
 are $0.006$ and $0.3$, and the parameter $\mu$ during inference is $0.5$.
For 5-task Split-UT-Zappos, $\lambda_1$, $\lambda_2$, $\lambda_3$, $\alpha$, $\beta$ and $\mu$ are set to $1.0$, $3 \times 10^{-6}$, $0.7$, $0.01$, $0.7$ and $0.02$, respectively. 
For 10-task Split-UT-Zappos, $\lambda_1$, $\lambda_2$, $\lambda_3$, $\alpha$, $\beta$ and $\mu$ are set to $0.5$, $10^{-7}$, $0.1$, $0.05$, $0.4$ and $0.03$.
We elaborate more details on hyper-parameter analysis in the appendix.



\subsection{Compared Baselines}
To demonstrate the effectiveness of the proposed method, we compare CompILer with state-of-the-art incremental learning methods, including prompt-free approaches~\cite{EWC, LwF, icarl} and prompt-based methods~\cite{L2P, Dual, CODA, Language_ICCV}. All the methods are rehearsal-free except iCaRL~\cite{icarl}. Note that, due to LGCL~\cite{Language_ICCV} relying on CLIP~\cite{CLIP} to achieve language guidance at the task level, it is limited by the length of class names per task. Thereby, LGCL fails to operate on the 5-task Split-UT-Zappos since the total length of class names exceeds the limitation.

To streamline our CompILer, we further implement a simplified version named \textbf{Sim-CompILer} and report its results. Sim-CompILer is optimized using cross entropy loss and is comprised solely of multi-pool prompt learning and generalized-mean prompt fusion. In other words, we exclude the object-injected prompting, directional decoupled loss, and reverse cross entropy loss, resulting in a large reduction of hyperparameters to only $\beta$, $\lambda_3$, and $\mu$.

\subsection{Comparison with the State-of-the-arts}
The compared results on Avg Acc and FTT are reported in Table~\ref{avg acc and ftt}. 
Overall, CompILer consistently outperforms all competitors on Avg Acc by a significant margin. 
For FTT scores, CompILer excels previous methods with 0.32\% on the 5-task Split-Clothing and with 0.14\% on the 5-task Split-UT-Zappos, while falling behind Dual-Prompt~\cite{Dual} and LGCL~\cite{Language_ICCV} for the 10-task Split-UT-Zappos. 
We notice that, the main reason is these methods sacrifice more plasticity for lower forgetting rates. 
Besides, the number of model parameters in these methods dynamically increases along with more incremental tasks arriving, whereas our CompILer does not rely on
imposing task-specific parameters to reduce the forgetting.

We also report the primitives accuracy and their HM in Table~\ref{tab: hm in clothing} and Table~\ref{tab: hm in ut}. 
Likewise, our method surpasses other methods considerably in terms of State and HM. 
Interestingly, the prompt-free methods~\cite{LwF,EWC,icarl} achieve higher accuracy in state prediction than object prediction for Split-Clothing, which is contrary to other results. 
This is because the states in Split-Clothing are color-related descriptions, which are easier to capture with the help of parameter fine-tuning. 
The prompt-based methods do not exhibit this phenomenon because their pre-trained backbones are initially trained for object classification, and are frozen across incremental sessions. 
As the performance improvements are mainly attributed to the accuracy of state recognition, it suggests that our model enhances the understanding on fine-grained compositionality.

\begin{table}[t]
\centering
\caption{State, Object and HM results on Split-UT-Zappos (5 tasks) and Split-UT-Zappos (10 tasks).}
\label{split-ut-zappos}
\resizebox{\textwidth}{!}{
\begin{tabular}{l||ccc||ccc}
\hline
Datasets & \multicolumn{3}{c||}{Split-UT-Zappos (5 tasks)} & \multicolumn{3}{c}{Split-UT-Zappos (10 tasks)} \\
\hline
Metrics & State & Object & HM & State & Object & HM \\
\hline
Upper Bound & 75.10±0.10 & 88.13±0.03 & 81.90±0.06 & 75.10±0.10 & 88.13±0.03 & 81.90±0.06 \\
\hdashline
EWC~\cite{EWC} & 47.95±1.26 & 76.53±0.91 & 58.90±0.53 & 39.29±2.69 & 67.64±1.97 & 49.69±2.30 \\
LwF~\cite{LwF} & 53.13±1.08 & 75.48±0.82 & 62.35±0.31 & 38.70±2.33 & 68.90±1.97 & 49.54±1.30 \\
iCaRL~\cite{icarl} & 51.71±0.95 & 75.03±0.49 & 61.22±0.78 & 38.94±2.01 & 67.10±1.05 & 49.27±1.58 \\
\hdashline
L2P~\cite{L2P} & 52.20±2.92 & 79.05±0.01 & 62.87±1.61 & 42.66±0.87 & 76.60±0.03 & 54.80±0.55 \\
Dual-Prompt~\cite{Dual} & 52.25±0.77 & 77.46±0.05 & 62.40±0.34 & 44.34±1.61 & 77.92±0.37 & 56.51±1.11 \\
LGCL~\cite{Language_ICCV} & - & - & - & 43.44±0.79 & \textbf{78.64}±0.64 & 55.96±0.43 \\
\hdashline
\textbf{Sim-CompILer} & 55.93±1.23 & \textbf{79.69}±0.06 & 65.72±0.53 & 45.88±0.38 & 75.72±0.67 & 57.14±0.06 \\
\textbf{CompILer} & \textbf{56.85}±0.34 & 79.56±0.04 & \textbf{66.31}±0.15 & \textbf{46.27}±1.56 & 76.65±1.19 & \textbf{57.69}±0.42 \\
\hline
\end{tabular}
}
\label{tab: hm in ut}
\end{table}

\begin{wraptable}{R}{0.58\textwidth}
\centering
\vspace{-0.35cm}
\caption{Ablation study on multi-pool prompt learning with Split-Clothing dataset.
}
\label{hierarchical prompt pool}
\begin{tabular}{ccc|ccc}
\hline
\multicolumn{3}{c|}{Prompt Pool}  & \multicolumn{3}{c}{Split-Clothing (5 tasks)}\\
\hline
C & S & O & Avg Acc & FTT(↓) & HM\\
\hline
\Checkmark &  &  & 80.22±0.41 & 14.23±0.44 & 88.85±0.16 \\
\Checkmark &  & \Checkmark & 88.10±0.11 & 7.79±0.04 & 93.55±0.04\\
\Checkmark & \Checkmark &  & 88.09±0.50 & \textbf{7.26}±0.54 & 93.52±0.13\\
\Checkmark & \Checkmark & \Checkmark & \textbf{88.38}±0.08 & 8.01±0.42 & \textbf{93.66}±0.02\\
\hline
\end{tabular}
\end{wraptable}
\subsection{Ablation Study and Analysis}

\textbf{Effect of multi-pool prompt learning.}
This experiment aims to delineate the contribution of three pools in CompILer. 
We firstly implement a baseline model with composition prompt pool only. 
Building upon the baseline, we develop two additional models,
which incorporate either object or state prompt pool.
As reported in Table~\ref{hierarchical prompt pool}, 
the inclusion of primitive prompt pool yields consistent gains over the baseline.
Furthermore, the best results are achieved when the model integrates all three pools simultaneously.
This experiment signifies the significant necessity of exploiting multiple prompt pools for composition-IL.


\begin{table*}[t]
\caption{
Ablative experiments for (a) object-injected state prompting, (b) prompt fusion method.
}
\begin{minipage}{.46\linewidth} 
  \centering
  \vspace{-0.2cm} 
  \subcaption{Object-injected state prompting.} 
    \vspace{-0.1cm}
    \begin{adjustbox}{width=1.05\linewidth}
    \begin{tabular}{c|ccc}
    \hline
    \multicolumn{1}{c|}{Dataset}  & \multicolumn{3}{c}{Split-Clothing (5 tasks)} \\
    \hline
    Metrics & Avg Acc & FTT(↓) & HM \\
    \hline
    None  & 88.45±0.10 & 7.93±0.11 & 93.70±0.03 \\
    S$\rightarrow$O & 88.27±0.02 & 7.99±0.05 & 93.67±0.01 \\
    O$\rightarrow$S  & \textbf{89.21}±0.24 & \textbf{7.26}±0.60 & \textbf{94.18}±0.06 \\
    \hline
    \end{tabular}
  \label{guidance}
  \end{adjustbox}
\end{minipage} %
\begin{minipage}{.55\linewidth} 
  \centering
  \vspace{-0.2cm} 
  \subcaption{Prompt fusion method.}
  \vspace{-0.1cm}
  \begin{adjustbox}{width=0.875\linewidth}
\begin{tabular}{c|ccc}
\hline
\multicolumn{1}{c|}{Dataset}  & \multicolumn{3}{c}{Split-Clothing (5 tasks)}\\
\hline
Metrics & Avg Acc & FTT(↓) & HM\\
\hline
Max  & 84.70±0.64 & 12.24±2.25 & 91.54±0.30\\
Mean & 87.80±0.12 & 7.82±0.01 & 93.38±0.03\\
GeM  & \textbf{89.21}±0.24 & \textbf{7.26}±0.60 & \textbf{94.18}±0.06\\
\hline
\end{tabular}
\end{adjustbox}
  \label{pooling}
\end{minipage} %
\label{tab:ablation}
\end{table*}

\textbf{Effect of object-injected state prompting.} 
To provide insights into object-injected state prompting, we compare three models: None (vanilla model), S$\rightarrow$O (state-injected object prompting) and O$\rightarrow$S (object-injected state prompting). As shown in Table~\ref{guidance}, compared to the None model, the S$\rightarrow$O exhibits a decrease in all metrics, implying that state prompts may interfere with the selection of object prompts. 
On the contrary, O$\rightarrow$S outperforms the None model as we expect.
This phenomenon validates our motivation that state recognition is harder than object recognition, and thereby the former cannot help the latter easily.
Yet, it is a promising direction for future research.


\textbf{Effect of generalized-mean prompt fusion.} 
This study aims to study the impact of prompt fusion on CompILer. 
As shown in Table~\ref{pooling}, 
GeM performs better than both max and mean pooling across various metrics.
It validates the benefit of GeM on mitigating irrelevant information in the selected prompts.
as it may hamper the model's attention on image tokens. 

\begin{table}[t]
\caption{Ablate the loss functions on Split-Clothing and Split-UT-Zappos.}
\label{loss}
\centering
\begin{tabular}{cccc|cc|cc}
\hline
\multicolumn{4}{c|}{Loss function}  & \multicolumn{2}{c|}{Split-Clothing (5 tasks)} & \multicolumn{2}{c}{Split-UT-Zappos (5 tasks)} \\
\hline
$\mathcal{L}_{CE}$ & $\mathcal{L}_{RCE}$ & $\mathcal{L}_{inter}$ & $\mathcal{L}_{intra}$ & Avg Acc & FTT(↓) & Avg Acc & FTT(↓)\\
\hline
\Checkmark & & & & 88.17±0.08 & 8.08±0.27 & 44.83±0.15 & 19.49±2.93\\
\Checkmark & \Checkmark & & & 88.36±0.37 & 8.33±0.11 & 45.47±0.07 & 20.14±0.43\\
\Checkmark & & \Checkmark & & 88.32±0.56 & 7.82±0.64 & 45.58±0.04 & 19.64±0.37\\
\Checkmark & & & \Checkmark & 88.42±0.30 & 8.23±0.06 & 45.62±0.13 & 20.13±0.14\\
\Checkmark & & \Checkmark & \Checkmark & 88.61±0.61 & 7.72±0.87 & 46.01±0.69 & 19.50±0.86\\
\Checkmark & \Checkmark & \Checkmark & \Checkmark & \textbf{89.21}±0.24 & \textbf{7.26}±0.60 & \textbf{46.48}±0.26 & \textbf{19.27}±0.75\\
\hline
\end{tabular}
\end{table}
\begin{figure*}[t]
  \centering
  \includegraphics[width=\textwidth]{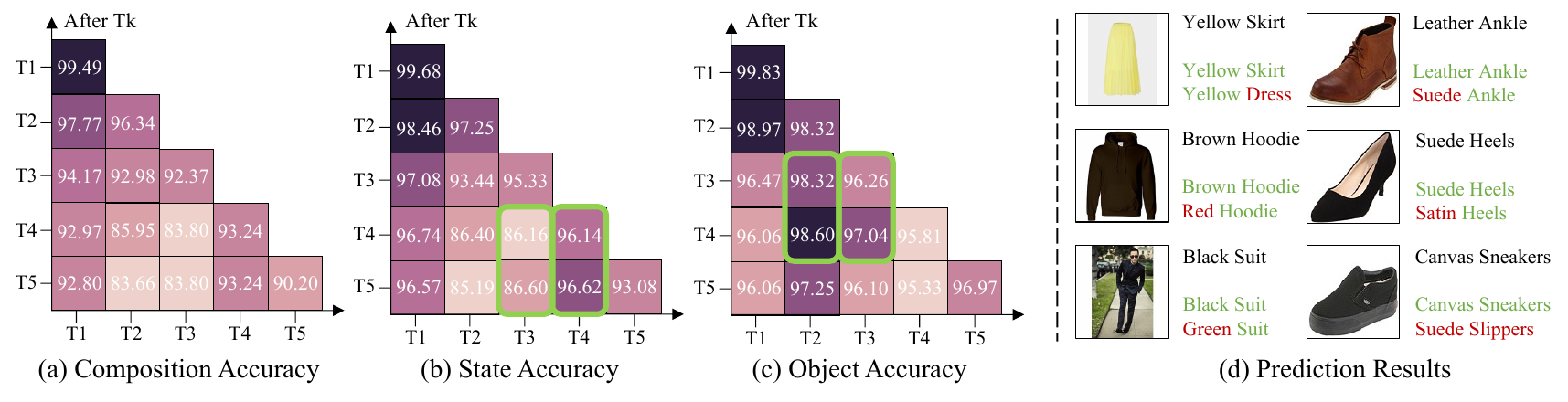}
   \vspace{-0.6cm}
   \caption{Results and analysis. (a) to (c) show accuracy of CompILer on composition, state, and object for each task in Split-Clothing. 
   The x-axis represents the test stream, and the y-axis denotes the status after training the $T_k$ task. Darker background color indicates higher accuracy. (d) displays some images and their predictions: top row is GT, middle row is CompILer prediction, and bottom row is L2P~\cite{L2P} prediction. \textcolor[RGB]{89,163,71}{Green} indicates \textcolor[RGB]{89,163,71}{correct} predictions, while \textcolor[RGB]{196,8,20}{red} indicates \textcolor[RGB]{196,8,20}{incorrect} predictions.}
   \label{fig:qualitative}
\end{figure*}


\textbf{Effect of loss functions.} 
As shown in Table~\ref{loss}, we investigate the influence of loss functions used in our model, 
including directional decoupled loss ($\mathcal{L}_{inter}$ and $\mathcal{L}_{intra}$) 
and symmetric cross entropy loss ($\mathcal{L}_{CE}$ and $\mathcal{L}_{RCE}$). 
The baseline model (the first row) includes all modules but is trained by cross entropy loss only. 
By adding the RCE loss, the model is equivalent to training with the SCE loss, 
which help to improve the robustness to noisy labels. 
The use of either $\mathcal{L}_{inter}$ or $\mathcal{L}_{intra}$ 
improves the performance on both datasets, and synchronously applying them 
witnesses all-around improvements compared to the baseline.
Eventually, we achieve the best results when combing all the loss terms during training.


\subsection{Additional Results and Analysis}
In order to study the repeatability characteristic in composition-IL,
we exhibit more results on Split-Clothing in Fig.~\ref{fig:qualitative}:
in (a), it shows a decreasing trend in composition accuracy along with the introduction of new tasks;
however, the green rectangles in (b) and (c) showcase that the accuracy occasionally increases as more tasks are learned.
We conjecture the reason is mostly attributed to the re-occurrence of primitive concepts. This forward transfer is critical for incremental learners. 
We compare the composition predictions between CompILer and L2P~\cite{L2P} in Fig.~\ref{fig:qualitative} (d). 
CompILer predicts all the images correctly, while L2P makes some mistakes, particularly for state labels.
This limitation arises from an excessive focus on the dominant object primitive, 
while weakening the attention toward state primitive. 
Fortunately, CompILer relieves the bias toward object classes, 
and enhances the perception on state classes.

\section{Conclusion}
In this paper, we have proposed a novel task coined compositional incremental learning (compostion-IL), which is stumbled by ambiguous composition boundary. 
To tackle it, we develop a learning-to-prompt model, namely CompILer.
Our model exploits multi-pool prompt learning to model composition and primitive concepts, object-injected state prompting to improve the selection of state prompts, and generalized-mean prompt fusion to eliminate irrelevant information. Extensive experiments on two tailored datasets show that CompILer achieves state-of-the-art performance. 
In the future, it is challenging yet potential to consider reasoning multiple state classes
per object.

\section{Acknowledgments}
This work was supported in part by the National Natural Science Foundation of China under Grant Numbers 62102061, 62272083 and 62472066, and in part by the Open Projects Program of State Key Laboratory of Multimodal Artificial Intelligence Systems.

\clearpage
\bibliographystyle{plain}
\bibliography{neurips_2024}

\begin{thebibliography}{10}

\bibitem{ViT_B16}
Alexey Dosovitskiy, Lucas Beyer, Alexander Kolesnikov, Dirk Weissenborn, Xiaohua Zhai, Thomas Unterthiner, Mostafa Dehghani, Matthias Minderer, Georg Heigold, Sylvain Gelly, et~al.
\newblock An image is worth 16x16 words: Transformers for image recognition at scale.
\newblock In {\em International Conference on Learning Representations}, 2020.

\bibitem{Consistent_Prompting}
Zhanxin Gao, Jun Cen, and Xiaobin Chang.
\newblock Consistent prompting for rehearsal-free continual learning.
\newblock In {\em IEEE Conf. Comput. Vis. Pattern Recog.}, 2024.

\bibitem{ADE}
Shaozhe Hao, Kai Han, and Kwan-Yee~K Wong.
\newblock Learning attention as disentangler for compositional zero-shot learning.
\newblock In {\em IEEE Conf. Comput. Vis. Pattern Recog.}, pages 15315--15324, 2023.

\bibitem{CognitiveScience1}
Geoffrey Hinton.
\newblock Some demonstrations of the effects of structural descriptions in mental imagery.
\newblock {\em Cognitive Science}, 3(3):231--250, 1979.

\bibitem{CZSL_AAAI_2024}
Chenchen Jing, Yukun Li, Hao Chen, and Chunhua Shen.
\newblock Retrieval-augmented primitive representations for compositional zero-shot learning.
\newblock In {\em Proceedings of the AAAI Conference on Artificial Intelligence}, pages 2652--2660, 2024.

\bibitem{IL_regular_1}
Sangwon Jung, Hongjoon Ahn, Sungmin Cha, and Taesup Moon.
\newblock Continual learning with node-importance based adaptive group sparse regularization.
\newblock {\em Adv. Neural Inform. Process. Syst.}, 33:3647--3658, 2020.

\bibitem{Language_ICCV}
Muhammad Gul Zain~Ali Khan, Muhammad~Ferjad Naeem, Luc~Van Gool, Didier Stricker, Federico Tombari, and Muhammad~Zeshan Afzal.
\newblock Introducing language guidance in prompt-based continual learning.
\newblock In {\em Int. Conf. Comput. Vis.}, pages 11429--11439, 2023.

\bibitem{CoT}
Hanjae Kim, Jiyoung Lee, Seongheon Park, and Kwanghoon Sohn.
\newblock Hierarchical visual primitive experts for compositional zero-shot learning.
\newblock In {\em Int. Conf. Comput. Vis.}, pages 5675--5685, 2023.

\bibitem{adam}
Diederik~P Kingma and Jimmy Ba.
\newblock Adam: A method for stochastic optimization.
\newblock {\em arXiv preprint arXiv:1412.6980}, 2014.

\bibitem{EWC}
James Kirkpatrick, Razvan Pascanu, Neil Rabinowitz, Joel Veness, Guillaume Desjardins, Andrei~A Rusu, Kieran Milan, John Quan, Tiago Ramalho, Agnieszka Grabska-Barwinska, et~al.
\newblock Overcoming catastrophic forgetting in neural networks.
\newblock {\em Proceedings of the national academy of sciences}, 114(13):3521--3526, 2017.

\bibitem{CLIB}
Hyunseo Koh, Dahyun Kim, Jung-Woo Ha, and Jonghyun Choi.
\newblock Online continual learning on class incremental blurry task configuration with anytime inference.
\newblock In {\em International Conference on Learning Representations}, 2022.

\bibitem{SCEN}
Xiangyu Li, Xu~Yang, Kun Wei, Cheng Deng, and Muli Yang.
\newblock Siamese contrastive embedding network for compositional zero-shot learning.
\newblock In {\em IEEE Conf. Comput. Vis. Pattern Recog.}, pages 9326--9335, 2022.

\bibitem{symnet}
Yong-Lu Li, Yue Xu, Xiaohan Mao, and Cewu Lu.
\newblock Symmetry and group in attribute-object compositions.
\newblock In {\em IEEE Conf. Comput. Vis. Pattern Recog.}, pages 11316--11325, 2020.

\bibitem{symnet-pami}
Yong-Lu Li, Yue Xu, Xinyu Xu, Xiaohan Mao, and Cewu Lu.
\newblock Learning single/multi-attribute of object with symmetry and group.
\newblock {\em IEEE Trans. Pattern Anal. Mach. Intell.}, 44(12):9043--9055, 2021.

\bibitem{CZSL_CVPR_2024}
Yun Li, Zhe Liu, Hang Chen, and Lina Yao.
\newblock Context-based and diversity-driven specificity in compositional zero-shot learning.
\newblock In {\em IEEE Conf. Comput. Vis. Pattern Recog.}, 2024.

\bibitem{LwF}
Zhizhong Li and Derek Hoiem.
\newblock Learning without forgetting.
\newblock {\em IEEE Trans. Pattern Anal. Mach. Intell.}, 40(12):2935--2947, 2017.

\bibitem{PIP}
Zilong Li, Yiming Lei, Chenglong Ma, Junping Zhang, and Hongming Shan.
\newblock Prompt-in-prompt learning for universal image restoration.
\newblock {\em arXiv preprint arXiv:2312.05038}, 2023.

\bibitem{CLC}
Weiduo Liao, Ying Wei, Mingchen Jiang, Qingfu Zhang, and Hisao Ishibuchi.
\newblock Does continual learning meet compositionality? new benchmarks and an evaluation framework.
\newblock {\em Adv. Neural Inform. Process. Syst.}, 2023.

\bibitem{PMGNet}
Yu~Liu, Jianghao Li, Yanyi Zhang, Qi~Jia, Weimin Wang, Nan Pu, and Nicu Sebe.
\newblock Pmgnet: Disentanglement and entanglement benefit mutually for compositional zero-shot learning.
\newblock {\em Computer Vision and Image Understanding}, 2024.

\bibitem{IL_replay_2}
David Lopez-Paz and Marc'Aurelio Ranzato.
\newblock Gradient episodic memory for continual learning.
\newblock {\em Adv. Neural Inform. Process. Syst.}, 30, 2017.

\bibitem{CompCos}
Massimiliano Mancini, Muhammad~Ferjad Naeem, Yongqin Xian, and Zeynep Akata.
\newblock Open world compositional zero-shot learning.
\newblock In {\em IEEE Conf. Comput. Vis. Pattern Recog.}, pages 5222--5230, 2021.

\bibitem{CL_survey_2}
Marc Masana, Xialei Liu, Bart{\l}omiej Twardowski, Mikel Menta, Andrew~D Bagdanov, and Joost Van De~Weijer.
\newblock Class-incremental learning: survey and performance evaluation on image classification.
\newblock {\em IEEE Trans. Pattern Anal. Mach. Intell.}, 45(5):5513--5533, 2022.

\bibitem{RedWine}
Ishan Misra, Abhinav Gupta, and Martial Hebert.
\newblock From red wine to red tomato: Composition with context.
\newblock In {\em IEEE Conf. Comput. Vis. Pattern Recog.}, pages 1792--1801, 2017.

\bibitem{MVP}
Jun-Yeong Moon, Keon-Hee Park, Jung~Uk Kim, and Gyeong-Moon Park.
\newblock Online class incremental learning on stochastic blurry task boundary via mask and visual prompt tuning.
\newblock In {\em IEEE Conf. Comput. Vis. Pattern Recog.}, pages 11731--11741, 2023.

\bibitem{CGE}
Muhammad~Ferjad Naeem, Yongqin Xian, Federico Tombari, and Zeynep Akata.
\newblock Learning graph embeddings for compositional zero-shot learning.
\newblock In {\em IEEE Conf. Comput. Vis. Pattern Recog.}, pages 953--962, 2021.

\bibitem{AttrAsOpe}
Tushar Nagarajan and Kristen Grauman.
\newblock Attributes as operators: factorizing unseen attribute-object compositions.
\newblock In {\em Eur. Conf. Comput. Vis.}, pages 169--185, 2018.

\bibitem{Task-driven}
Senthil Purushwalkam, Maximilian Nickel, Abhinav Gupta, and Marc'Aurelio Ranzato.
\newblock Task-driven modular networks for zero-shot compositional learning.
\newblock In {\em Int. Conf. Comput. Vis.}, pages 3593--3602, 2019.

\bibitem{Prompt_Gradient_Projectio}
Jingyang Qiao, Xin Tan, Chengwei Chen, Yanyun Qu, Yong Peng, Yuan Xie, et~al.
\newblock Prompt gradient projection for continual learning.
\newblock In {\em The Twelfth International Conference on Learning Representations}, 2023.

\bibitem{GeM}
Filip Radenovi{\'c}, Giorgos Tolias, and Ond{\v{r}}ej Chum.
\newblock Fine-tuning cnn image retrieval with no human annotation.
\newblock {\em IEEE transactions on pattern analysis and machine intelligence}, 41(7):1655--1668, 2018.

\bibitem{CLIP}
Alec Radford, Jong~Wook Kim, Chris Hallacy, Aditya Ramesh, Gabriel Goh, Sandhini Agarwal, Girish Sastry, Amanda Askell, Pamela Mishkin, Jack Clark, Gretchen Krueger, and Ilya Sutskever.
\newblock Learning transferable visual models from natural language supervision.
\newblock In {\em International Conference on Machine Learning}, pages 8748--8763, 2021.

\bibitem{CognitiveScience2}
Pasko Rakic, Jean-Pierre Bourgeois, and Patricia~S Goldman-Rakic.
\newblock {\em The self-organizing brain: from growth cones to functional networks}.
\newblock Elsevier, 1994.

\bibitem{icarl}
Sylvestre-Alvise Rebuffi, Alexander Kolesnikov, Georg Sperl, and Christoph~H Lampert.
\newblock icarl: Incremental classifier and representation learning.
\newblock In {\em IEEE Conf. Comput. Vis. Pattern Recog.}, pages 2001--2010, 2017.

\bibitem{CODA}
James~Seale Smith, Leonid Karlinsky, Vyshnavi Gutta, Paola Cascante{-}Bonilla, Donghyun Kim, Assaf Arbelle, Rameswar Panda, Rog{\'{e}}rio Feris, and Zsolt Kira.
\newblock Coda-prompt: Continual decomposed attention-based prompting for rehearsal-free continual learning.
\newblock In {\em IEEE Conf. Comput. Vis. Pattern Recog.}, pages 11909--11919, 2023.

\bibitem{IL_distill_1}
Filip Szatkowski, Mateusz Pyla, Marcin Przewi{\k{e}}{\'z}likowski, Sebastian Cygert, Bart{\l}omiej Twardowski, and Tomasz Trzci{\'n}ski.
\newblock Adapt your teacher: Improving knowledge distillation for exemplar-free continual learning.
\newblock In {\em Proceedings of the IEEE/CVF Winter Conference on Applications of Computer Vision}, pages 1977--1987, 2024.

\bibitem{NotMeetStrongPT}
Yu-Ming Tang, Yi-Xing Peng, and Wei-Shi Zheng.
\newblock When prompt-based incremental learning does not meet strong pretraining.
\newblock In {\em IEEE Conf. Comput. Vis. Pattern Recog.}, pages 1706--1716, 2023.

\bibitem{HiDe}
Liyuan Wang, Jingyi Xie, Xingxing Zhang, Mingyi Huang, Hang Su, and Jun Zhu.
\newblock Hierarchical decomposition of prompt-based continual learning: Rethinking obscured sub-optimality.
\newblock {\em Adv. Neural Inform. Process. Syst.}, 36, 2024.

\bibitem{CL_survey_1}
Liyuan Wang, Xingxing Zhang, Hang Su, and Jun Zhu.
\newblock A comprehensive survey of continual learning: Theory, method and application.
\newblock {\em IEEE Trans. Pattern Anal. Mach. Intell.}, 2024.

\bibitem{CANet}
Qingsheng Wang, Lingqiao Liu, Chenchen Jing, Hao Chen, Guoqiang Liang, Peng Wang, and Chunhua Shen.
\newblock Learning conditional attributes for compositional zero-shot learning.
\newblock In {\em IEEE Conf. Comput. Vis. Pattern Recog.}, pages 11197--11206, 2023.

\bibitem{IL_regular_2}
Wenjin Wang, Yunqing Hu, Qianglong Chen, and Yin Zhang.
\newblock Task difficulty aware parameter allocation \& regularization for lifelong learning.
\newblock In {\em IEEE Conf. Comput. Vis. Pattern Recog.}, pages 7776--7785, 2023.

\bibitem{s-prompt}
Yabin Wang, Zhiwu Huang, and Xiaopeng Hong.
\newblock S-prompts learning with pre-trained transformers: An occam’s razor for domain incremental learning.
\newblock {\em Adv. Neural Inform. Process. Syst.}, 35:5682--5695, 2022.

\bibitem{SCE}
Yisen Wang, Xingjun Ma, Zaiyi Chen, Yuan Luo, Jinfeng Yi, and James Bailey.
\newblock Symmetric cross entropy for robust learning with noisy labels.
\newblock In {\em Int. Conf. Comput. Vis.}, pages 322--330, 2019.

\bibitem{Dual}
Zifeng Wang, Zizhao Zhang, Sayna Ebrahimi, Ruoxi Sun, Han Zhang, Chen{-}Yu Lee, Xiaoqi Ren, Guolong Su, Vincent Perot, Jennifer~G. Dy, and Tomas Pfister.
\newblock Dualprompt: Complementary prompting for rehearsal-free continual learning.
\newblock In {\em Eur. Conf. Comput. Vis.}, pages 631--648, 2022.

\bibitem{L2P}
Zifeng Wang, Zizhao Zhang, Chen{-}Yu Lee, Han Zhang, Ruoxi Sun, Xiaoqi Ren, Guolong Su, Vincent Perot, Jennifer~G. Dy, and Tomas Pfister.
\newblock Learning to prompt for continual learning.
\newblock In {\em IEEE Conf. Comput. Vis. Pattern Recog.}, pages 139--149, 2022.

\bibitem{ut-zappos}
Aron Yu and Kristen Grauman.
\newblock Fine-grained visual comparisons with local learning.
\newblock In {\em IEEE Conf. Comput. Vis. Pattern Recog.}, pages 192--199, 2014.

\bibitem{CognitiveScience3}
Alan~L Yuille and Chenxi Liu.
\newblock Deep nets: What have they ever done for vision?
\newblock {\em Int. J. Comput. Vis.}, 129(3):781--802, 2021.

\bibitem{IVR}
Tian Zhang, Kongming Liang, Ruoyi Du, Xian Sun, Zhanyu Ma, and Jun Guo.
\newblock Learning invariant visual representations for compositional zero-shot learning.
\newblock In {\em Eur. Conf. Comput. Vis.}, pages 339--355, 2022.

\bibitem{CSCNet}
Yanyi Zhang, Qi~Jia, Xin Fan, Yu~Liu, and Ran He.
\newblock Cscnet: Class-specified cascaded network for compositional zero-shot learning.
\newblock In {\em IEEE International Conference on Acoustics, Speech and Signal Processing}, pages 3705--3709, 2024.

\bibitem{IL_replay_1}
Yaqian Zhang, Bernhard Pfahringer, Eibe Frank, Albert Bifet, Nick Jin~Sean Lim, and Yunzhe Jia.
\newblock A simple but strong baseline for online continual learning: Repeated augmented rehearsal.
\newblock {\em Adv. Neural Inform. Process. Syst.}, 35:14771--14783, 2022.

\end{thebibliography}

\end{document}